\documentclass[runningheads]{llncs}  

\usepackage{graphicx}
\usepackage{orcidlink}  
\usepackage{amsmath, amssymb, amsfonts}
\usepackage{mathtools}
\usepackage{mathrsfs}
\usepackage{graphicx}
\usepackage{graphics}
\usepackage{adjustbox}
\usepackage[font=small]{caption}
\usepackage{subcaption}
\usepackage{enumitem}
\usepackage{multirow}
\usepackage{booktabs}
\usepackage{flushend}
\usepackage{xcolor}
\usepackage{textcomp}
\usepackage{manyfoot}
\usepackage{arydshln}
\usepackage[title]{appendix}
\usepackage[capitalise]{cleveref}
\usepackage[
    round-mode=places, detect-weight=true,
    round-pad=false, exponent-product=\cdot,
    retain-unity-mantissa=false]{siunitx}
\usepackage{algorithm}
\usepackage{algorithmicx}
\usepackage{algpseudocode}
\usepackage{listings}
\usepackage{courier}
\usepackage{times}
\usepackage{helvet}
\usepackage{authorindex}

\usepackage[backend=biber,
  style=lncs
]{biblatex}
\title{Leap: Inductive Link Prediction via Learnable Topology Augmentation}
\titlerunning{Leap: Inductive Link Prediction}  

\author{
    Ahmed E. Samy\orcidlink{0000-0002-5392-6531} \and
    Zekarias T. Kefato\orcidlink{0000-0001-7898-0879} \and
    \v{S}ar\={u}nas Girdzijauskas\orcidlink{0000-0003-4516-7317}
}

\authorrunning{A. E. Samy et al.}  

\institute{
    KTH Royal Institute of Technology, Stockholm, Sweden \\
    \email{aesy@kth.se, zekarias@kth.se, sarunasg@kth.se}
}

\begin{document}
 
 \maketitle              

\begin{abstract}
Link prediction is a crucial task in many downstream applications of graph machine learning. To this end, Graph Neural Network (GNN) is a widely used technique for link prediction, mainly in transductive settings, where the goal is to predict missing links between existing nodes. However, many real-life applications require an inductive setting that accommodates for new nodes, coming into an existing graph. Thus, recently inductive link prediction has attracted considerable attention, and a multi-layer perceptron (MLP) is the popular choice of most studies to learn node representations. However, these approaches have limited expressivity and do not fully capture the graph's structural signal. Therefore, in this work we propose LEAP, an inductive link prediction method based on LEArnable toPology augmentation. Unlike previous methods, LEAP models the inductive bias from both the structure and node features, and hence is more expressive. To the best of our knowledge, this is the first attempt to provide structural contexts for new nodes via learnable augmentation in inductive settings. Extensive experiments on seven real-world homogeneous and heterogeneous graphs demonstrates that LEAP significantly surpasses SOTA methods. The improvements are  up to 22\% and 17\% in terms of AUC and average precision, respectively. The code and datasets are available on GitHub\footnote{\url{https://github.com/AhmedESamy/LEAP/}}.
  
\end{abstract}
\section{Introduction}
\begin{figure*}
\centering
\includegraphics[width=1.05\textwidth, height = 7.5cm]{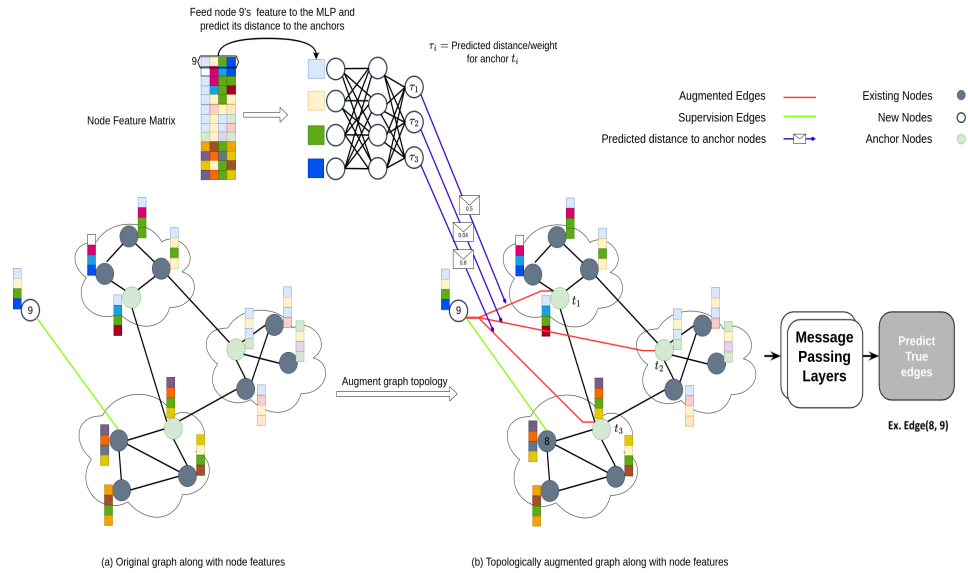}
\caption{An overview of LEAP; anchors per community are sampled (Fig1(a)). The red edges (Fig1(b)) signify the topology augmentation as predicted by the MLP. Each augmented/predicted edge is weighted by an activation from the final layer of the MLP and links a given newcomer node with an anchor node.
The green edge is the inductive link to predict.} \label{intro_fig}
\end{figure*}

Link prediction serves as a fundamental task in numerous machine learning applications based on graphs, such as predicting co-authorship connections within citation networks, and interactions between genes and proteins within biological networks~\cite{ref1}~\cite{ref2}~\cite{ref3}. There are two variants of link prediction: \textit{transductive} link prediction, which predicts links between observed nodes in the graph, and \textit{inductive} link prediction, which is used for newly-coming, often isolated nodes. Inductive link prediction in graphs can be categorized into semi-inductive or fully-inductive approaches. Semi-inductive methods involve predicting links between nodes that have been observed previously and those that are unseen. On the other hand, fully inductive methods entail predicting links solely among unseen nodes, such as predicting connections within an entirely new graph utilizing a model trained on a prior one. Recently, semi-inductive link prediction has gained considerable attention due to its necessity in most real-life applications with the cold-start problem, where there is minimum information available about the newcomer nodes. For example, predicting how new users should be connected with other existing users to improve the performance of recommender systems. With that said, this paper aims to tackle the cold-start problem within the framework of inductive link prediction.

A common approach to link prediction is through node embeddings, where nodes are mapped to low-dimensional vectors, and links are predicted according to the similarity/proximity of nodes in the embedding space ~\cite{ref4}~\cite{ref5}. Graph Neural Networks (GNNs) is a common approach for node embedding and link prediction ~\cite{ref4}~\cite{ref6}~\cite{ref7}. GNNs are message-passing architectures, where messages (i.e., node features) are passed along the edges and aggregated at every node to compute its embedding. Although GNNs are highly expressive, it is hard (if not impossible) to apply existing architectures for inductive link prediction where the topology information about the target nodes is missing. 

In pursuit of this objective, numerous approaches have adopted Multi-Layer Perceptron (MLP)-based encoders for inductive link prediction. DEAL~\cite{deal} incorporates node features and structure during training, yet suffers from a shallow structure encoder, hence constraining expressiveness \cite{graph2feat}. G2G~\cite{g2g} introduces a deep encoder heavily reliant on node features, resulting in challenges in distinguishing nodes with similar features \cite{deal}. A recent technique, Graph2Feat \cite{graph2feat}, employs knowledge distillation to transfer insights from a GNN teacher to an MLP student for link prediction. However, during inference, both G2G and Graph2Feat disregard structural information in semi-inductive settings, thereby lacking the desired expressiveness compared to GNNs that could leverage graph topology if accessible \cite{ref8}.

While recent research has been exploring learning representations for paths \cite{zhu2021neural} and local subgraphs \cite{zhang2017weisfeiler, zhang2018link} to facilitate inductive link prediction, these approaches operate under the assumption that newcomer nodes are already connected during inference, contrary to the settings presumed in this paper. In practical contexts, nodes frequently join a large-scale graph sequentially, initially lacking any neighbors \cite{liu2020heterogeneous}.

To utilize graph topology during inference, even when such is scarce, we introduce LEAP (\textbf{LEA}rning to\textbf{P}ology). LEAP starts by selecting a set of anchor nodes in the graph where different anchor selection methods based on structural properties such as PageRank or centrality measure \cite{pagerank} are explored. Next, we augment the input graph by assigning new, \textit{weighted} connections  between newly-comer nodes and the anchor nodes. This augmentation enables newly-comer nodes to develop tailored topological connections and take advantage of the graph connectivity (see Figure~\ref{intro_fig}b). Afterwards, LEAP utilizes message-passing (MP) layers, including GNN, that use the learned topology augmentation to create meaningful representations for both new and existing nodes in the augmented graph, as depicted in Figure \ref{intro_fig}.

LEAP is applicable to both homogeneous and heterogeneous graphs and outperforms state of the art on  a range of datasets, reaching up to 22\% and 17\%, 10.6\% and 16.1\% improvements on homogeneous and heterogeneous graphs, on the area under the ROC curve, and average precision, respectively. 

To summarize, our contributions are stated as follows:
\begin{itemize}
 \item We devise LEAP, an \textit{expressive} approach that enables message passing between the new nodes and the anchors (in the exiting graph) via learnable topological augmentation. \textit{To the best of our knowledge, this is the first attempt to provide structural contexts for the new nodes via learnable augmentation under inductive settings.}
 
\item We conduct extensive experiments on seven real-world homogeneous and heterogeneous graphs for transductive and inductive link predictions. Our experiments demonstrate the significant improvements of LEAP over SOTAs approaches.

\end{itemize}

\section{Related Work}
There is a plethora of approaches for predicting links in different types of graphs. The most common setting is transductive, although there has also been much research on (fully) inductive link prediction for knowledge graphs in particular as GNNs are inherently inductive. However, knowledge graphs and heterogeneous graphs differ in a slight but distinct way. In knowledge graphs, the nodes, or entities, can have multiple node types simultaneously, whereas heterogeneous graphs can only have one type at any given time.  

Many approaches introduced to date utilize MLP-based encoders to compute node embeddings \cite{g2g, deal, graph2feat}. DEAL \cite{deal} uses an MLP-based encoder to compute a node embedding based on node feature information, and a linear encoder to compute a node embedding for the structural information. An alignment mechanism is then used to combine these two encoders and produce a single node embedding. However, DEAL may not possess the same expressiveness as our proposed method, as MLPs are less expressive than GNNs. On the other hand, G2G \cite{g2g} maps nodes to a density to allow for uncertainty in prediction tasks; however, this encoder only creates a node embedding based on node features. The structural information is only incorporated through a ranking approach incorporated into the loss function. Another recent method, Graph2Feat \cite{graph2feat}, uses knowledge distillation from a teacher GNN model to a student MLP model. However, Graph2Feat also generalizes this method to perform on both heterogeneous and homogeneous graphs. Graph2Feat trains the teacher GNN model first, and during inference time, the method acts as an MLP encoder. Unlike a GNN-based method such as LEAP, these methods may lose some expressiveness during the inference time.

Other research directions have undertaken the task of acquiring representations for paths \cite{zhu2021neural} and local subgraphs \cite{zhang2017weisfeiler, zhang2018link} in order to facilitate inductive link prediction. These approaches make an assumption that the newcomer nodes are already interconnected during the inference phase, which is not an assumption of LEAP. In practical scenarios, nodes often come in on a sequential basis to a large graph without any neighbors \cite{liu2020heterogeneous}, which is a scenario where new, incoming nodes are not interconnected.

\section{Preliminaries}
In the following, we provide definitions for the necessary notation used through out the paper and the basic elements of our model.

\textbf{Graph.} We define a generic undirected graph $G = (V, E, \phi, \psi, \mathbf{X})$ with a set of nodes $V = \lbrace 1, ..., N \rbrace $ and a set of edges $E = \lbrace e_i = (u, v) | u, v \in V, i = 1, ..., M \rbrace $. For every node $u \in V$, the function $\phi: V \rightarrow \mathcal{A}$ maps $u$ to its type $\phi(u) = u_t \in \mathcal{A}$, and for every edge $e = (u, v) \in E$, the function $\psi: E \rightarrow \mathcal{R}$ maps $e$ to its edge type $\psi(e) = e_t \in \mathcal{R}$.
Where $\mathcal{A}$ and $\mathcal{R}$ define the set of all possible node types and edge types, respectively.
The matrix $\mathbf{X} \in \mathbb{R}^{N \times d}$ is a feature matrix and that characterizes every node $u$ using its feature vector $\mathbf{x}_u$.
Denoted $N_u$, where $N_u = \lbrace (u, v) \in E \rbrace$ defines the neighorhood of $u$.

A \textit{homogeneous} graph is a graph, where $\vert \mathcal{A} \mathcal \vert = 1$ and $ \vert \mathcal{R} \vert = 1$; and a \textit{heterogeneous} graph is a graph, where $|\mathcal{A}|+|\mathcal{R}|>2$.


We use the widely adopted encoder-decoder architecture, where the encoder is a message passing neural network (MPNN), also known as GNN.
A GNN works by successively applying $L$ message passing layers to refine the feature vectors $\mathbf{x}_u$ for all $u \in V$ as follows
\vspace{0.1cm}
\[
\mathbf{h}^{l + 1}_u = UPDATE(\mathbf{W}_1^l\mathbf{h}_u^l, AGGREGATE(\lbrace \mathbf{W}_2^l\mathbf{h}_v^l \vert v \in N_u \rbrace))
\]

Where, $\mathbf{W^l_*}$ is a parameter of the model, and $\mathbf{h}^l_u$ is the intermediate representation of node $u$, with $\mathbf{h}^0_u = \mathbf{x}_u$. AGGREGATE is a permutation invariant, and usually a non-parametric aggregation function (e.g., sum and mean) that aggregate messages $(\mathbf{W}^l_2\mathbf{h}_v)$ from neighbors of $u$, $N_u$. The UPDATE function, which is commonly parameterized by a neural network and followed by an activation function, updates node $u$'s state by using its own previous state $(\mathbf{W}_1^l\mathbf{h}_u^l)$ and the aggregated message from its neighbors.

The final representation $\mathbf{Z}^{(L)} \in \mathbb{R}^{N \times d}$ or simply $\mathbf{Z}$,  is then used in the decoder. For simplicity, we assume the GNN output and input features has the same dimension, $d$. The decoder is an edge scoring function, which we shall define in Section~\ref{sec:encodign_decoding}.


\section{LEAP}

Although  Graph Neural Networks (GNN) are expressive enough to handle node-centered and edge-centered tasks, they can not operate under an inductive setting, where newly-coming nodes lack connection to the existing graph \cite{hamilton2017inductive, deal, graph2feat}. 
Partly, this is due to their dependency on the topology of the graph to propagate messages\cite{brody2021attentive, kipf2016semi, zeng2019graphsaint}.
Several efforts \cite{deal, g2g, graph2feat} have tackled this issue by integrating a Multi-Layer Perceptron (MLP) as part of the encoder that only needs MLP for inference.
Despite providing fast inference, MLPs have less inductive bias compared to GNNs.

For this reason, we propose LEAP, a MLP-GNN based framework for inductive link prediction. LEAP improves performance on link prediction by carefully bringing MLP and GNN together, which in turn trade of between expressiveness and inference time. That is, we lift-up GNNs with MLP so that a GNN can also handle new nodes with no connection to the graph. Consequently, using a GNN both during training and inference time, enabling us to exploit GNNs expressive power with an empirically insignificant decrease in inference speed.
Our design is general and flexible enough, and can be applied on both homogeneous and heterogeneous graphs.

Inductive learning in LEAP is mainly facilitated by the use of anchors, this is inline with existing studies for knowledge graph embedding \cite{galkin2021nodepiece}. The following section provides a salient detail on the selection strategy.



 \subsection{Anchor Selection}
\label{sec:4.2}
To choose anchors, one intuitive way is to select top-k nodes with the highest similarities to each new-comer node. However, this is a less scalable heuristic and do not provide guarantee that the chosen anchors are optimal for predicting missing links. We rather opt for \textit{learning} links to a fixed set of anchors, which are sampled either randomly or based on structural measures such as degree or PageRank \cite{pagerank} scores. These learned links (augmentations) allow new nodes to be wired to the graph topology with adaptive weights. Intuitively, these weights can be seen as learned distances (similarities) to the anchors. 

Thus, we sample a set $A = \lbrace a_i, ..., a_k \rbrace$ of $k$ anchor nodes using one of the following three techniques.\\ 
\textbf{Random}: anchors are sampled uniformly at random out of $N$ nodes in the graph without replacement as an unordered set of $k$ nodes.\\ \textbf{Deterministic}: anchors are an ordered set of nodes selected according to centrality measures such as the degree or PageRank \cite{pagerank}.\\
\textbf{Community-based}: first communities are identified using Louvain clustering method \cite{louvain}, then from each community, anchors are sampled using one of the aforementioned methods.

In our experiments, degree-based anchor selection performs better than randomly sampling the anchors on homogeneous graphs, and comparably well to the community-based method.

After that, given a new-node $i$, our goal is to predict its closeness score $\tilde{w}_{i, j}$ to each anchor node $a_j \in A$. Let $\hat{E} = \lbrace \tilde{w}_{i, j} \rbrace$ be the set of all pairwise prediction scores between every new-node $i$ and anchor $a_j$, then we propagate Messages over $E \cup \hat{E}$. 
The above steps are carried out by the MLP-GNN based encoder discussed below. 

 \subsection{Encoding and Decoding}
 \label{sec:encodign_decoding}
 Once anchors are selected, encoding is carried out in to steps, which are \emph{linking} and \emph{message passing}, using an MLP-Linker and a GNN.
 
 \textbf{Linking}. 
It is carried out using an MLP-Linker that predicts the closeness score $\tilde{w}_{i, j}$ between a certain new node $i$ and each anchor $a_j \in A$.
 The linker, ${g: \mathbb{R}^{d} \rightarrow [0, 1]^k}$, takes as input node features and produces a $k$--dimensional output.
 Consider $\mathbf{x}_i$, then $\mathbf{\tilde{w}}_i = g(\mathbf{x}_i)$, corresponds to the predicted score vector to the $k$ anchors. The $j$--th component, $\tilde{w}_{i, j}$, of $\mathbf{\tilde{w}}_i$ is the predicted score to anchor $a_j \in A$.
 Given $\hat{E} = \lbrace \tilde{w}_{i, j} \rbrace$, we then generate an augmented graph $G' = (V, E')$, where ${E' = E \cup \hat{E}}$.

 We materialize $g$ as shown in Eq.~\ref{eq:hom_mlp_linker} and Eq.~\ref{eq:het_mlp_linker}, for homogeneous and heterogeneous graphs, respectively.
 \begin{equation}\label{eq:hom_mlp_linker}
     \mathbf{\tilde{w}}_i = g(\mathbf{x}_i) = \sigma(\mathbf{x_i}\mathbf{W}^T + \mathbf{b})
 \end{equation}
 \begin{equation}\label{eq:het_mlp_linker}
     \mathbf{\tilde{w}}_i = g(\mathbf{x}_i) = \sigma(\mathbf{x_i}\mathbf{W}_{t}^T + \mathbf{b}_{t})
 \end{equation}
 where, $\sigma$ is a sigmoid activation function, and $\mathbf{W}, \mathbf{b}, \mathbf{W_t}$ and $\mathbf{b_t}$ are model parameters. For heterogeneous graphs, we use type specific parameters, $\mathbf{W_t}$ and $\mathbf{b_t}$, for every $t \in \mathcal{A}$ and $\phi(i) = t$.

 Intuitively, the linker model provides knowledge of how the new nodes should be connected to the input graph, thus making an informed choice of the anchors via the \textit{learned weights}.
  
 \textbf{Message Passing}. 
 Given an augmented graph $G'$, now we can apply a GNN that can accommodate new nodes and learn their representations by passing messages over the augmented and existing edges.
 Formally, the $l$--th layer of GNN with $L$ message passing layers is given as
 
 \begin{equation}\label{eq:gnn_layer}
\mathbf{h}^l_i = \texttt{ReLU} (\mathbf{W}_1^l\cdot \mathbf{h}^{l-1}_{i} + \sum_{j \in N(i)} \alpha_{ij} \cdot  \mathbf{W}_2^l\mathbf{h}^{l-1}_{j}) 
\end{equation}

where, $\alpha_{i,j}$ is a learnable or precomputed edge weight for the edge $(i, j)$, depending on the choice of the GNN architecture, e.g. GAT\cite{brody2021attentive}, GCN\cite{kipf2016semi}.
The weights $\mathbf{W}_1^l$ and $\mathbf{W}_2^l$ are parameters of the GNN, and $\mathbf{h}_i^l$ is the intermediate state of node $i$ at the $l$--th layer.
The full GNN is formed by stacking $L$ such message passing layers to obtain the final representation $\mathbf{Z}$.  

We treat the predicted edges as a special type of edge, akin to heterogeneous edges, and hence we use two GNNs, one that work over $E$ and another over $\hat{E}$, and their respective outputs are $\mathbf{Z}_E$ and $\mathbf{Z}_{\hat{E}}$. $\mathbf{Z}$ is an aggregation of the two.

The homogeneous layer formulation in Eq.~\ref{eq:gnn_layer} can be trivially extended to a heterogeneous layer by simply using edge type specific $\mathbf{W}_{r}^l$, for $r \in \mathcal{R}$ transformation function, instead of $\mathbf{W}_2^l$.

\textbf{Decoding}. After computing $\mathbf{Z}$, the remaining task is to decode each edge from $\mathbf{Z}$.
 There are different option for decoding, e.g., using an MLP, dot product, and cosine similarity. In this study we simply use dot product, and decode each edge $(i, j)$ using their respective representation $\mathbf{z}_i$ and $\mathbf{z}_j$ as:
 \[
    \texttt{dec}(\mathbf{z}_i,  \mathbf{z}_j) = \mathbf{z}_i^T \mathbf{z}_j
\]

 \subsection{Training}
 To train LEAP, we sample a set $I \subset V$ of $n$ inductive nodes, and compute their closeness (similarity) to all the anchor nodes.
 That is, for each inductive node $i$, we compute
\begin{equation}
    w_{i,j} = \frac{1}{dist(i, j)}
\end{equation}
which is, its closeness to anchor $a_j \in A$, because $dist(i, j)$ is the shortest distance between $i$ and $j$ in $G$. Thus, $\mathbf{w}_i$ captures $i$'s closeness to all the anchors.
After materializing $\mathbf{w}_i$ for every $i \in I$, we remove them from the graph and hold them out for training LEAP.
 
 LEAP is trained end-to-end to jointly optimize the parameters of both the MLP-linker and GNN.
 To this end, it trains using a loss functions that has two respective terms.
 First, we define the loss term pertinent to the parameres of the MLP as
 \begin{equation}\label{eq:mlp_loss_term}
     \mathcal{L}_{\texttt{MLP}} = \vert \vert \Tilde{\mathbf{w}}_{i} - \mathbf{w}_{j} \vert \vert ^2
 \end{equation}
and then, the second term with respect to the paramters of the GNN is defined by
 \begin{equation}\label{eq:}
     \mathcal{L}_{\texttt{GNN}} = \log \sigma(\texttt{dec}(\mathbf{z}_i,  \mathbf{z}_j)) + \sum \mathbb{E}_{p \sim \mathcal{N}_i} \big [ \log \sigma (\texttt{dec}(\mathbf{z}_i, -\mathbf{z}_p)) \big]
 \end{equation}
 where $\mathcal{N}$ is the noise distribution and $\mathcal{N}_i$ are $q$ negative samples for node $i$.
Finally, the full loss function is given by
\begin{equation}
    \mathcal{L} = \mathcal{L}_{\texttt{GNN}} + \gamma \mathcal{L}_{\texttt{MLP}}
\end{equation}
As $\mathcal{L}$ jointly optimize the model parameters of the GNN and MLP, LEAP can learn closeness of new nodes to the fixed set of anchors from their features. 

\section{Experiments}
\subsection{Datasets}

To perform link prediction, we evaluate on seven real-world datasets. For homogeneous graphs, we use Wikipedia (chameleon and crocodile entities), Twitch (a followship graph of gamers) \cite{wikipedia}, and PubMed (a citation graph of papers for the biomedical literature) \cite{pubmed}. We have ACM, DBLP (academic graphs), and IMDB \cite{heterodata} for heterogeneous graphs. 

\subsection{Baseline Methods}
For transductive experiments, we have random-walk-based methods such as DeepWalk \cite{deepwalk} and Metapath2Vec \cite{metapath2vec}. Also, we include Graph Convolution Network (GCN)\cite{kipf2016semi}, Variational Graph Auto-Encoder (VGAE) \cite{gae}  and  Graph Attention Network (GAT2) \cite{brody2021attentive} as GNN models. For inductive link prediction baselines, we include the state-of-the-art methods: DEAL \cite{deal}, Graph2Gauss \cite{g2g}, and Graph2Feat \cite{graph2feat}. For metapath2vec, we choose the same metapaths reported in \cite{wang2019heterogeneous}. Additionally, we add following metapaths: paper-term-paper on ACM, paper-author-paper For DBLP and director-movie-keyword-movie-director on IMDB.

\subsection{Experimental Setup}
\begin{table}[ht!]
\centering
\caption{Evaluation - graph datasets.}
\begin{adjustbox}{width=0.6\textwidth}
\small
\setlength{\tabcolsep}{0.5em}

\begin{tabular}{lccccc
} 
\toprule
\textbf{Dataset} &\textbf{\# nodes} & \textbf{\# edges} & \textbf{Node types} & \textbf{Edge types}\\
\midrule
Wiki (Chameleon) & \num{2277} & \num{62792} & \num{1}&\num{1}\\ 
Wiki (Crocodile)  & \num{11631} & \num{341691} & \num{1}&\num{1}\\ 
PubMed       & \num{19717} & \num{88648} & \num{1}& \num{1}\\ 
Twitch   & \num{7126} & \num{77774}  & \num{1}& \num{1}\\ 
ACM           & \num{10942} & \num{547872}&\num{4}    & \num{8}\\ 
DBLP   &   \num{26128} & \num{239566} &\num{4}    & \num{6}\\ 
IMDB       & \num{21420} & \num{86642} &\num{4}    & \num{6}\\ 

\bottomrule
\end{tabular}
\label{tab:1}
\end{adjustbox}
\end{table}
\vspace{-0.1cm}

\textbf{Inductive setting} In this setting, the nodes in the validation set are not seen in the training set. Similarly, the nodes in the testing set are newly coming nodes that are not included in the training or validation datasets. We evaluate inductive link prediction between the new and existing nodes. For all the datasets, we randomly assign 80/10/10 \% of the nodes for training/validation/testing. For ACM and DBLP, the newly-coming nodes are of node type "paper", while for IMDB, we choose nodes from the node type "movie" as the newly coming nodes. \\

\textbf{Transductive setting } 

In the transductive setting, we assign 80/10/10 \% of the edges to training/validation/testing respectively. However, as link prediction is easier for Wikipedia (Crocodile entity) than other datasets, we split the edges according to a 10/45/45 \% split. \\

The evaluation results are reported on the testing set for the best-performing hyperparameters. We tune the hyperparameters for all the methods on the same validation set of each dataset. For fair comparisons, we report the average of five runs for all methods, and the hidden size is set to 128. The evaluation metrics are the area under the ROC curve (AUC) and the average precision (AP). 
\subsection{Inductive Link Prediction}
\begin{table*}[ht!]
\centering
\caption{Inductive Link Prediction - Homogeneous Graphs.}
\begin{adjustbox}{width=1\textwidth}
\small
\begin{tabular}{lcccccccc}
\toprule
\textbf{Approach} & \multicolumn{2}{c}{\textbf{Wikipedia (Chameleon)}} & \multicolumn{2}{c}{\textbf{Wikipedia (Crocodile)}} &\multicolumn{2}{c}{\textbf{PubMed}}&\multicolumn{2}{c}{\textbf{Twitch}}\\
\cmidrule(lr){2-3}\cmidrule(lr){4-5}\cmidrule(lr){6-7}\cmidrule(lr){8-9}
 & AUC\%& AP~\% & AUC\%& AP~\% & AUC\%& AP~\% & AUC\%& AP~\%\\
\midrule
MLP& 60.7$\pm$1.5	&64.3$\pm$0.3	&	93.1$\pm$0.1&	92.8$\pm$0.1	&	90.0$\pm$0.1	&89.0$\pm$0.2	&	81.1$\pm$0.4	&81.4$\pm$0.2\\
DEAL&

65.2$\pm$2.6	&65.2$\pm$1.6	&	91.1$\pm$0.2&	89.4$\pm$0.1	&	91.1$\pm$0.1&	89.1$\pm$0.2	&	83.1$\pm$0.2&	83.6$\pm$0.1
\\
G2G&

60.0$\pm$0.2	&63.9$\pm$0.1	&	93.5$\pm$0.1&	95.0$\pm$0.1	&	91.1$\pm$0.9	&92.0$\pm$0.9	&	78.5$\pm$0.1&	82.1$\pm$0.0\\
Graph2Feat&

72.5$\pm$0.7&	71.0$\pm$0.4	&	92.9$\pm$0.8	&91.6$\pm$1.2	&	92.8$\pm$0.1&	92.0$\pm$0.1	&	82.0$\pm$0.1&	82.3$\pm$0.1\\
LEAP \small(no augment)&

71.0$\pm$0.5	&65.4$\pm$0.4	&	71.6$\pm$0.8&	66.5$\pm$0.1	&	90.8$\pm$0.0&	88.9$\pm$0.1	&	50.1$\pm$0.1&	48.1$\pm$0.1
\\
LEAP \small(unweighted augment)&

88.7$\pm$0.1	&81.0$\pm$0.1	&	95.0$\pm$0.2&	93.2$\pm$0.5	&	94.7$\pm$0.0	&90.2$\pm$0.0	&	94.5$\pm$0.2	&90.2$\pm$0.2
\\
LEAP \small(learned augment)&

\textbf{89.5}$\pm$\textbf{0.1}	&\textbf{81.9}$\pm$\textbf{0.1}	&	\textbf{96.5}$\pm$\textbf{0.1}	&\textbf{95.4}$\pm$\textbf{0.0}	&	\textbf{94.9}$\pm$\textbf{0.6}	&\textbf{92.8}$\pm$\textbf{0.3}	&	\textbf{95.4}$\pm$\textbf{0.0}	&\textbf{91.6}$\pm$\textbf{0.2}\\
\bottomrule
\end{tabular}
\label{tab:3}
\end{adjustbox}
\end{table*}

\begin{table*}[ht!]
\centering
\caption{Inductive Link Prediction - Heterogeneous Graphs.}
\begin{adjustbox}{width=1\textwidth}
\small
\setlength{\tabcolsep}{0.5em}
\begin{tabular}{lcccccc}
\toprule
\textbf{Approach} & \multicolumn{2}{c}{\textbf{ACM}} & \multicolumn{2}{c}{\textbf{DBLP}} &\multicolumn{2}{c}{\textbf{IMDB}}\\
\cmidrule(lr){2-3}\cmidrule(lr){4-5}\cmidrule(lr){6-7}
 & AUC\%& AP~\% & AUC\%& AP~\% & AUC\%& AP~\% \\
\midrule
MLP&
76.8$\pm$1.6 &	75.3$\pm$1.7	&	79.2$\pm$0.4&	77.9$\pm$0.2	&	69.1$\pm$0.6&	66.1$\pm$1.0 
\\
DEAL&
54.6$\pm$1.7	&52.6$\pm$1.7	&	52.5$\pm$0.6	&55.2$\pm$0.5	&	57.7$\pm$0.9&	57.0$\pm$0.8
\\
G2G&
54.0$\pm$0.3	&54.6$\pm$0.3	&	54.1$\pm$0.1	&51.1$\pm$0.1	&	54.6$\pm$0.6&	54.1$\pm$0.2

\\
Graph2Feat&
76.8$\pm$0.8	&75.3$\pm$0.9	&	79.3$\pm$0.1	&78.1$\pm$0.1	&	69.1$\pm$2.0	&66.9$\pm$2.5
\\
LEAP \small(no augmented edges)&
71.4$\pm$1.1&	70.0$\pm$2.4	&	72.4$\pm$1.0	&68.0$\pm$1.2	&	70.8$\pm$0.5	&64.7$\pm$0.6
\\
LEAP \small(learned weighted augment)&
\textbf{89.7}$\pm$\textbf{1.5}	&\textbf{87.5}$\pm$\textbf{2.0}	&	\textbf{86.7}$\pm$\textbf{0.9}	&\textbf{86.1}$\pm$\textbf{1.0}	&	\textbf{76.8}$\pm$\textbf{1.3}&	\textbf{74.3}$\pm$\textbf{1.6}
 \\
\bottomrule
\end{tabular}
\label{tab:4}
\end{adjustbox}
\end{table*}

The comparative results of the introduced methods on inductive link prediction are listed in Tables \ref{tab:3} and \ref{tab:4}. On the one hand, GCN and random-based methods such as DeepWalk and Metapath2Vec are among our baselines. However, these methods require knowing the neighborhood structure around each node, which is unavailable in the inductive settings. Therefore, we exclude those methods from this part of the experiments. On the other hand, MLP, G2G, and DEAL are inherently inductive link prediction methods. Their results show that they perform better on Wikipedia (Crocodile) and PubMed than on Wikipedia (Chameleon) and Twitch datasets. We hypothesize that these methods rely more on the features than the graph's structure. Hence they perform better on the datasets where the node features are possibly more vital, such as PubMed, which has 500 node features. Graph2Feat has better modeling for the graph's structure via knowledge distillation. That possibly explains why Graph2Feat performs better on Twitch and Wikipedia (Chameleon). However, knowledge distillation, acting as regularization, seems less beneficial on the other datasets.

For heterogeneous graphs, G2G and DEAL are not originally designed to operate on them, hence the poor performance. To the best of our knowledge, except for Graph2Feat, there needs to be more research in inductive link prediction for heterogeneous graphs. Although many methods exist for inductive link prediction on knowledge graphs, reasoning over them is a different problem; for example, entities (nodes) in knowledge graphs may appear as different types; node "Barack Obama" can be of president and husband node types at the same time \cite{graph2feat}.

The LEAP method, as shown in Tables \ref{tab:3} and \ref{tab:4}, shows significant improvements on all datasets, particularly on the heterogeneous ones. Concisely, LEAP reaches up to 22\% and 17\%,  10.6\% and 16.1\% improvements on homogeneous and heterogeneous graphs, on AUC and average precision, respectively. Our conclusions are through augmentation, LEAP attains greater expressiveness and achieves better performance on inductive link prediction. LEAP surpasses the state-of-the-art for heterogeneous graphs and shows to be a general approach that achieves new state-of-the-art results on different types of graphs, such as homogeneous, heterogeneous, directed and undirected. Augmented edges and sampling anchors are essential for LEAP's success, as shown in Table \ref{tab:3}.
\vspace{-0.2cm}

\subsection{Transductive Link Prediction}
\begin{table*}[ht!]
\centering
\caption{Transductive Link Prediction - Homogeneous Graphs.}
\begin{adjustbox}{width=1\textwidth}
\small
\setlength{\tabcolsep}{0.5em}
\begin{tabular}{lcccccccc}
\toprule
\textbf{Approach} & \multicolumn{2}{c}{\textbf{Wikipedia (Chameleon)}} & \multicolumn{2}{c}{\textbf{Wikipedia (Crocodile)}} &\multicolumn{2}{c}{\textbf{PubMed}}&\multicolumn{2}{c}{\textbf{Twitch}}\\
\cmidrule(lr){2-3}\cmidrule(lr){4-5}\cmidrule(lr){6-7}\cmidrule(lr){8-9}
 & AUC\%& AP~\% & AUC\%& AP~\% & AUC\%& AP~\% & AUC\%& AP~\%\\
\midrule
MLP & 91.5$\pm$0.1 &	91.6$\pm$0.0	 &	95.2$\pm$0.1 &	94.5$\pm$0.0	&	94.1$\pm$0.2	& 93.8$\pm$0.2	&	82.7$\pm$0.2	& 83.7$\pm$0.1\\
DEAL&
90.8$\pm$0.1 &	89.6$\pm$0.1	&	90.4$\pm$0.1&	89.3$\pm$0.2		&90.7$\pm$0.1	&88.7$\pm$0.1	&	82.5$\pm$0.7	&83.0$\pm$0.3\\
G2G&
93.2$\pm$0.0 & 93.6$\pm$0.1	&	94.3$\pm$0.4	&95.3$\pm$0.2	&	94.6$\pm$0.0 &	94.8$\pm$0.1 &		82.7$\pm$0.2 &
82.8$\pm$0.7
\\
DeepWalk&
94.8$\pm$0.2	&95.6$\pm$0.3	&	93.1$\pm$0.0	&95.6$\pm$0.0	&	95.0$\pm$0.1&	96.9$\pm$0.1	&	67.7$\pm$0.3&	75.8$\pm$0.2\\
GCN&
94.4$\pm$0.1&	94.4$\pm$0.1	&	95.8$\pm$0.1	&96.3$\pm$0.1		&97.5$\pm$0.0	&97.4$\pm$0.1		&73.5$\pm$0.2&	73.9$\pm$0.3
\\
GIN&
95.7$\pm$0.2&	96.3$\pm$0.2	&	96.3$\pm$0.0&	97.3$\pm$0.1	&	95.5$\pm$0.1	&95.8$\pm$0.1	&	81.4$\pm$0.6	&83.1$\pm$0.5
\\
GAT2&
96.4$\pm$0.1	&96.7$\pm$0.1	&	95.9$\pm$0.1&	96.4$\pm$0.0	&97.6$\pm$0.0&	97.6$\pm$0.0	&	75.1$\pm$0.3	&75.6$\pm$0.4
\\
VGAE&
96.1$\pm$0.1&	97.3$\pm$0.1	&	76.8$\pm$0.3	&80.5$\pm$0.3&		\textbf{98.7}$\pm$\textbf{0.1}&	97.7$\pm$0.1	&	76.8$\pm$0.3&	80.5$\pm$0.2
\\
Graph2Feat&
95.9$\pm$0.1	&96.7$\pm$0.1	&	95.6$\pm$0.0&	95.0$\pm$0.0	&	96.1$\pm$0.0&	96.1$\pm$0.0	&	83.0$\pm$0.0&	83.9$\pm$0.1
\\
LEAP&
\textbf{98.1}$\pm$\textbf{0.1}	&\textbf{98.3}$\pm$\textbf{0.1}		&\textbf{98.0}$\pm$\textbf{0.0}&	\textbf{98.1}$\pm$\textbf{0.0}	&	98.4$\pm$0.0&	98.7$\pm$0.0&		\textbf{84.5}$\pm$\textbf{0.1}	&\textbf{85.9}$\pm$\textbf{0.1}
\\
\bottomrule
\end{tabular}
\end{adjustbox}
\label{tab:5}
\end{table*}

\begin{table*}[ht!]
\centering
\caption{Transductive Link Prediction - Heterogeneous Graphs.}
\begin{adjustbox}{width=1\textwidth}
\small
\setlength{\tabcolsep}{1em}

\begin{tabular}{lcccccc}
\toprule
\textbf{Approach} & \multicolumn{2}{c}{\textbf{ACM}} & \multicolumn{2}{c}{\textbf{DBLP}} &\multicolumn{2}{c}{\textbf{IMDB}}\\
\cmidrule(lr){2-3}\cmidrule(lr){4-5}\cmidrule(lr){6-7}
 & AUC\%& AP~\% & AUC\%& AP~\% & AUC\%& AP~\% \\
\midrule
MLP
&
86.7$\pm$3.2&	85.3$\pm$3.5	&	90.3$\pm$3.0&	88.9$\pm$2.9	&	88.6$\pm$0.4&	86.1$\pm$0.4\\
DEAL&
72.0$\pm$0.2&	72.4$\pm$0.1	&	72.7$\pm$0.1&	74.1$\pm$0.8	&	77.2$\pm$0.4	&78.7$\pm$0.4\\
G2G&
68.1$\pm$0.2&	66.5$\pm$0.1	&	73.9$\pm$0.5&	72.9$\pm$0.3	&	64.2$\pm$0.5	&61.5$\pm$0.3
\\
Metapath2Vec&
80.1$\pm$0.8&	84.1$\pm$0.6	&	70.9$\pm$0.2&	70.9$\pm$0.2	&	60.7$\pm$2.2	&61.9$\pm$2.5
\\
Graph2Feat&
93.6$\pm$0.5&	93.0$\pm$0.4	&	91.5$\pm$0.1&	90.3$\pm$0.1&		88.6$\pm$0.3	&86.1$\pm$0.1\\
LEAP&
\textbf{95.4}$\pm$\textbf{0.1}&	\textbf{94.9}$\pm$\textbf{0.1}	&	\textbf{97.2}$\pm$\textbf{0.0}&	\textbf{97.0}$\pm$\textbf{0.0}	&	\textbf{94.0}$\pm$\textbf{0.1}	&\textbf{93.3}$\pm$\textbf{0.1}
\\
\bottomrule
\end{tabular}
\end{adjustbox}

\label{tab:6}
\end{table*}

The results of transductive link prediction are shown in Tables \ref{tab:5} and \ref{tab:6}. As the structure information is available  for all methods under transductive settings, the methods which are more effective in capturing both features and structures, can surpass the others. However, for some datasets, the features can provide more signal information than the structure such as Twitch. DeepWalk seems to exhibit good performance across the datasets, except for Twitch. Based on the previous analysis, DeepWalk lacks access to the features, which possibly explains the lower performance on Twitch. Although GCN have access to both the features and structure, DEAL and G2G seem more efficient in incorporating the feature information. Therefore, they perform better especially on Twitch.

For the heterogeneous datasets, as mentioned before, in inductive settings, G2G and DEAL lack access, by design, to the heterogeneity knowledge in those graphs; this is why they perform poorly on the heterogeneous graphs. Contrastly, Graph2Feat is well-designed for the latter settings and thus outperforms the other baselines. As for Metapath2Vec, despite being designed to run on heterogeneous graphs, the method is highly conditioned on the choice of the meta-paths, which requires, in turn, domain knowledge. Furthermore, Metapath2Vec lacks access to the node features, similar to DeepWalk on homogeneous link prediction.

Viewing the results in Tables \ref{tab:5} and \ref{tab:6}, LEAP demonstrates consistently superior performance over the baselines across the datasets. As the topological information is available in transductive settings, the augmentation seems to be less viable in that context. We conclude that the encoder part of LEAP has more expressiveness, that it is able to efficiently encode both feature and structure information, therefore, the enhanced performance. Furthermore, considering the heterogeneous knowledge (i.e., node and edge types) results in high-quality embeddings and better performance. 

\subsection{How does selecting the anchors affect LEAP?}
\vspace{-0.7cm}

\begin{table}[ht!]
\centering
\caption{Link prediction results using different anchor selection strategies.}
\begin{adjustbox}{width=0.6\textwidth}
\small
\setlength{\tabcolsep}{0.5em}
\begin{tabular}{lcccccc}
\toprule
\textbf{Dataset} & \multicolumn{2}{c}{\textbf{Random}} & \multicolumn{2}{c}{\textbf{Degree Centrality}} & \multicolumn{2}{c}{\textbf{Community}}\\
\cmidrule(lr){2-3}\cmidrule(lr){4-5}\cmidrule(lr){6-7}
 & AUC\%& AP~\% & AUC\%& AP~\%& AUC\%& AP~\%\\
\midrule
Chameleon
& 84.8&	77.2 & \textbf{88.9}&	\textbf{81.2}&		\textbf{88.9}&	80.2\\
Twitch&
92.4&	89.4	&	\textbf{94.3}&	\textbf{91.3}&		93.9&	91.2\\
\bottomrule
\end{tabular}
\end{adjustbox}
\label{tab:7}
\end{table}

We experimented on homogeneous graphs for the proposed methods in section \ref{sec:4.2} to choose the anchors. The results of each method on inductive link predictions are listed in Table \ref{tab:7}. We have uniformly random anchor selection, degree-based, and community-based anchor selection. For the degree-based method, we select the most important nodes with the highest degree. For the community-based, we cluster the graph using the Louvain algorithm \cite{louvain} and select the nodes with the highest degrees within each community.

We hypothesize that considering the graph topology leads to a smarter choice of anchors. Viewing the results in Table \ref{tab:7}, selecting anchors based on highest degrees gives the best performance. Our analysis is that combining high-degree anchors with learnable edge weights offers more tailored insight into the graph's topology than uniform sampling (low-degrees) anchors. As shown in Table \ref{tab:7}, selecting anchors within each community performs similarly to the degree-based method. We hypothesize that this is because enough global high degree anchor nodes sufficiently represent each community; thus, community-based anchor selection adds no benefit to the performance. We use degree-based anchor selection since this provides similar results and is less computationally expensive than the community-based method. On heterogeneous graphs, we choose anchors uniformly from all node types. Based on our experiments, uniformly sampling anchors from each node type seems sufficient for heterogeneous graphs.

\label{section: 5.6}
\subsection{Hyperparameter sensitivity}
\subsubsection{Number of Inputs and Anchors}
\begin{figure*}
\centering
\begin{tabular}{cc}
  \includegraphics[width=0.4\linewidth]{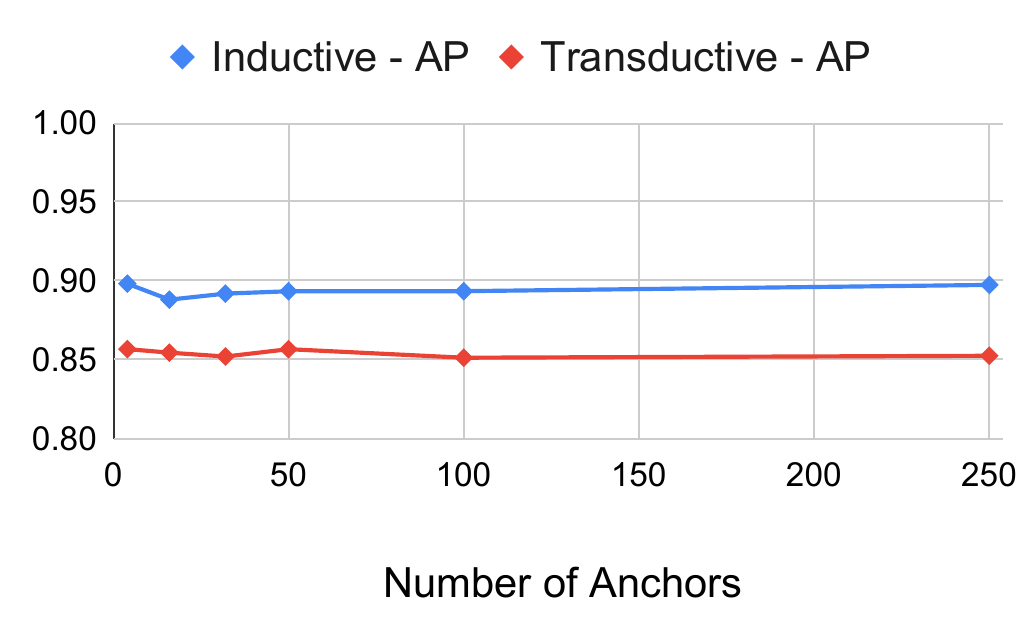} &   \includegraphics[width=0.4\linewidth]{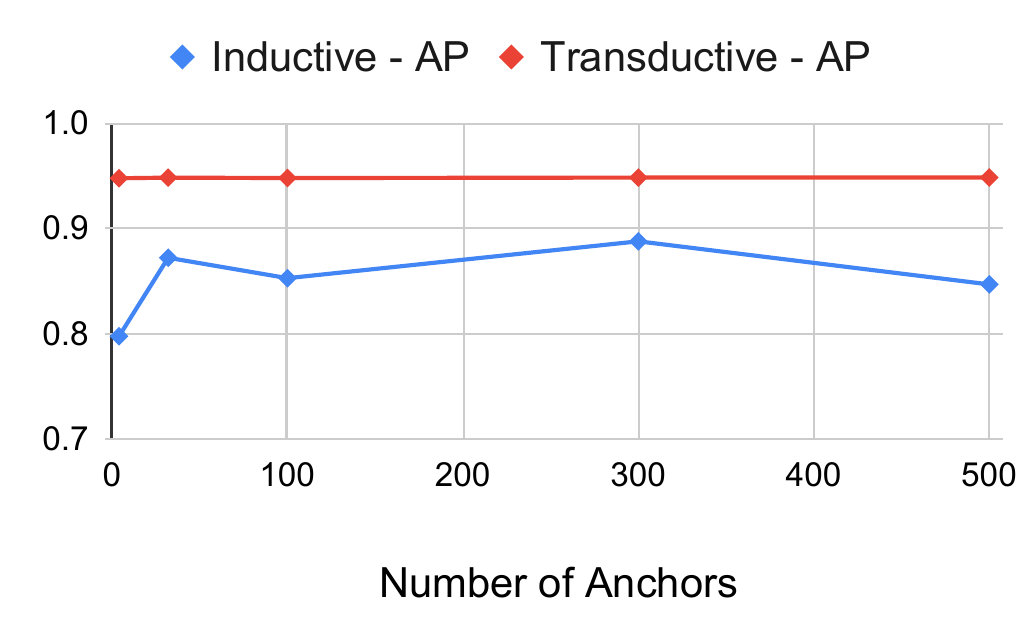} \\
(a) Twitch Anchor & (b) ACM Anchor \\[6pt]
  \includegraphics[width=0.4\linewidth]{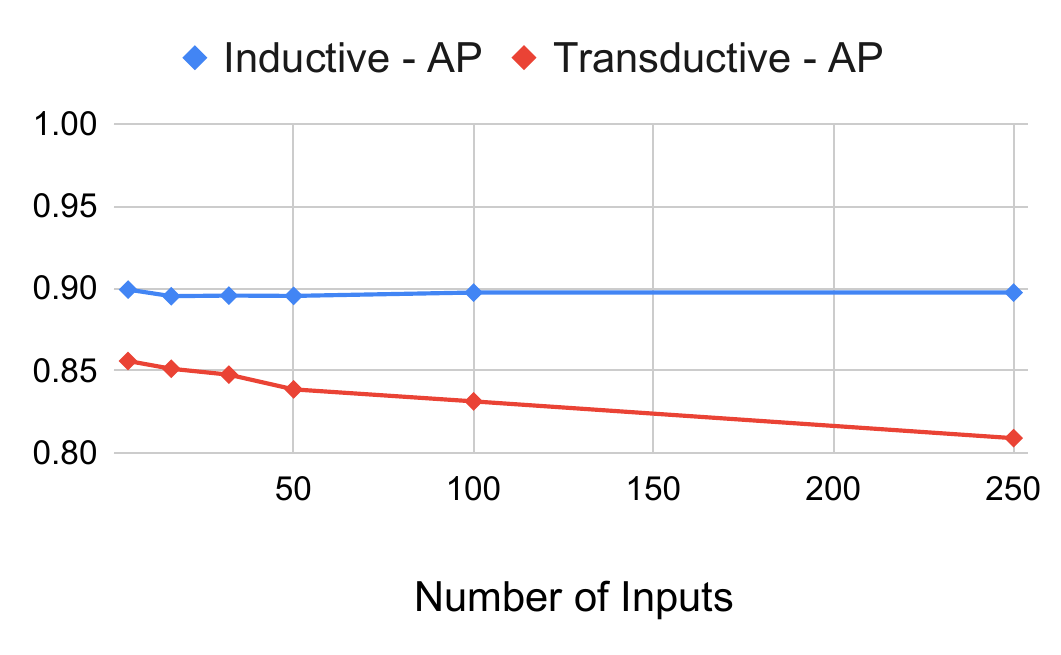} &   \includegraphics[width=0.4\linewidth]{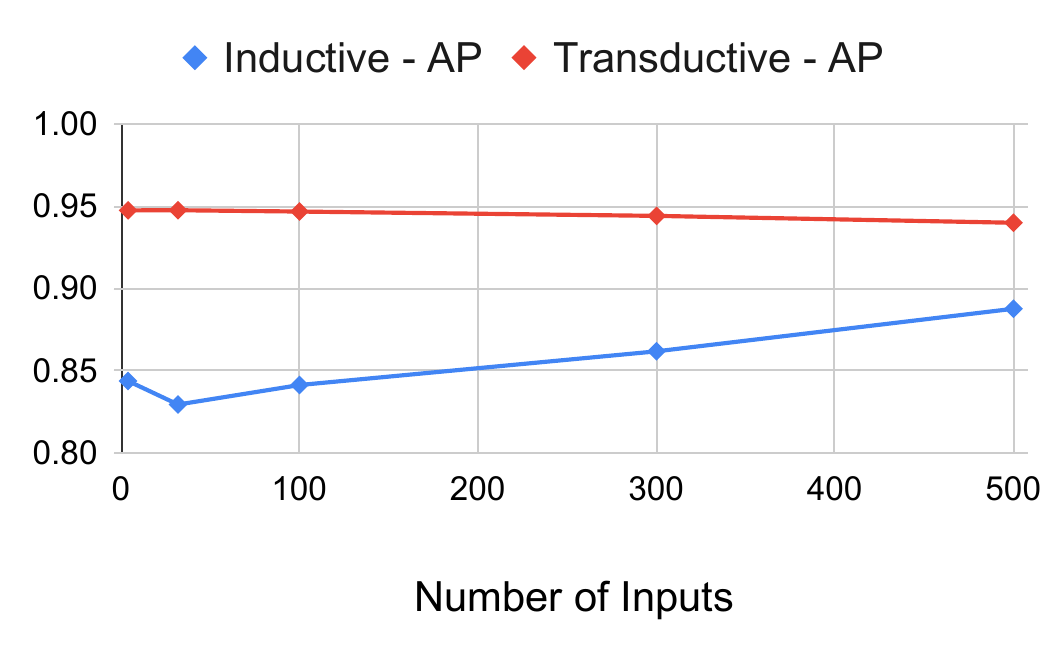} \\
(c) Twitch Input & (d) ACM Input \\[6pt]
\end{tabular}
\caption{Anchor \& Input Ablations}
\label{fig:comb_anch}
\end{figure*}

The performance of LEAP, in both inductive and transductive settings, is evaluated depending on varying numbers of input and anchor nodes. We select one homogeneous graph, Twitch, and one heterogeneous graph, ACM, to perform the study.


We examined the impact of varying the number of anchor nodes for Twitch and ACM. For Twitch, we experimented with the values of $[4, 16, 32, 50, 100, 250]$, while for ACM, we used $[4, 32, 100, 300, 500]$. In the inductive and transductive settings for Twitch, the number of anchors did not significantly affect the AP. Likewise, in the transductive setting for ACM, the number of anchors did not affect the AP. However, in the inductive setting, the AP varied with the number of anchors. This variation could be attributed to the complexity of heterogeneous graphs. As a result, more anchors are required to capture the graph topology adequately. We observed an increase in AP from 4 anchors to 300 anchors, but the addition of more anchors did not appear to enhance the model's performance beyond 300 anchors.


We also investigated the model's performance with respect to the number of inputs, as shown in Figure \ref{fig:comb_anch}. In the inductive setting for Twitch, the AP did not change with the number of inputs. However, in the transductive setting, there is a decrease in the AP. Upon analysis, this could potentially be due to the underlying structural properties of Twitch when in transductive settings. 
Examining Twitch dataset in the transductive case, we have observed much more nodes that have a very low degree. Consequently, when the number of inputs increases, we likely sampled more low-degree nodes, which added noise rather than structural information to the model. For ACM, there is no effect on AP in the transductive case. However, in the inductive case, the AP increases with the number of inputs. Similar to the effect of increasing the number of anchors, this could be attributed to the complex structure of heterogeneous datasets. An increase in the number of inputs is essentially equivalent to an increase in the number of training examples, which resulted in improved performance.



\subsubsection{Gamma coefficient}

The next hyperparameter we evaluate is gamma: the coefficient in the MLP loss function. For this ablation study, we only evaluate the homogeneous dataset, Twitch. We evaluate the performance of the method in regards to AP when varying gamma across the following values: $[0.001, 0.005, 0.05, 0.5, 1]$. In both the transductive and inductive settings, gamma does not materially affect the results, and our model is robust compared to the coefficient for the MLP loss function. This implies that learning in an end-to-end fashion through the main loss function is sufficient.

\section{Conclusion}
The existing approaches for inductive link prediction are essentially MLP encoders, primarily designed for homogeneous graphs. Hence, they do not consider the structural contexts around the new nodes and have less expressiveness, i.e., less inductive bias in their models compared to GNNs. Motivated by the latter, we dedicated this work to bringing the GNN's expressiveness into inductive link prediction.
Thus, we proposed LEAP, an encoder that incorporates learnable augmentation of the input graph by learning edge weights between anchors nodes selected from the graph and new nodes. We conducted extensive experiments in which LEAP exhibited superior performance over the SOTA methods and demonstrated a general approach that can be applied to homogeneous and heterogeneous graphs, under both inductive and transductive settings. Furthermore, we carried out comprehensive ablation study to understand how the hyperparameters impact the performance of LEAP. Our findings are the number of inputs and anchors as well as the hidden size are crucial factors for the performance of LEAP in inductive settings. Our future work involve extending LEAP to knowledge graphs and investigating more sophisticated ways of sampling anchors, such as Gumbel-Softmax sampling proposed in \cite{gumbelsoftmax}. 

%
%
%
%






\printbibliography

@article{ref1,
  title={Content-based citation recommendation},
  author={Bhagavatula, Chandra and Feldman, Sergey and Power, Russell and Ammar, Waleed},
  journal={arXiv preprint arXiv:1802.08301},
  year={2018}
}

@inproceedings{liu2020heterogeneous,
  title={A heterogeneous graph neural model for cold-start recommendation},
  author={Liu, Siwei and Ounis, Iadh and Macdonald, Craig and Meng, Zaiqiao},
  booktitle={Proceedings of the 43rd international ACM SIGIR conference on research and development in information retrieval},
  pages={2029--2032},
  year={2020}
}

@inproceedings{ref2,
  title={Graph trend filtering networks for recommendation},
  author={Fan, Wenqi and Liu, Xiaorui and Jin, Wei and Zhao, Xiangyu and Tang, Jiliang and Li, Qing},
  booktitle={Proceedings of the 45th International ACM SIGIR Conference on Research and Development in Information Retrieval},
  pages={112--121},
  year={2022}
}

@article{ref3,
  title={Simple embedding for link prediction in knowledge graphs},
  author={Kazemi, Seyed Mehran and Poole, David},
  journal={Advances in neural information processing systems},
  volume={31},
  year={2018}
}

@article{ref4,
  title={Inductive representation learning on large graphs},
  author={Hamilton, Will and Ying, Zhitao and Leskovec, Jure},
  journal={Advances in neural information processing systems},
  volume={30},
  year={2017}
}

@inproceedings{ref5,
  title={SchemaWalk: Schema Aware Random Walks for Heterogeneous Graph Embedding},
  author={Samy, Ahmed E and Giaretta, Lodovico and Kefato, Zekarias T and Girdzijauskas, {\v{S}}ar{\=u}nas},
  booktitle={Companion Proceedings of the Web Conference 2022},
  pages={1157--1166},
  year={2022}
}

@article{ref6,
  title={Pitfalls of graph neural network evaluation},
  author={Shchur, Oleksandr and Mumme, Maximilian and Bojchevski, Aleksandar and G{\"u}nnemann, Stephan},
  journal={arXiv preprint arXiv:1811.05868},
  year={2018}
}

@article{ref7,
  title={Link prediction based on graph neural networks},
  author={Zhang, Muhan and Chen, Yixin},
  journal={Advances in neural information processing systems},
  volume={31},
  year={2018}
}

@article{g2g,
  title={Deep gaussian embedding of graphs: Unsupervised inductive learning via ranking},
  author={Bojchevski, Aleksandar and G{\"u}nnemann, Stephan},
  journal={arXiv preprint arXiv:1707.03815},
  year={2017}
}

@article{deal,
  title={Inductive link prediction for nodes having only attribute information},
  author={Hao, Yu and Cao, Xin and Fang, Yixiang and Xie, Xike and Wang, Sibo},
  journal={arXiv preprint arXiv:2007.08053},
  year={2020}
}

@article{ref8,
  title={Graph-less neural networks: Teaching old mlps new tricks via distillation},
  author={Zhang, Shichang and Liu, Yozen and Sun, Yizhou and Shah, Neil},
  journal={arXiv preprint arXiv:2110.08727},
  year={2021}
}

@article{pagerank,
  title={The PageRank citation ranking: bringing order to the web},
  author={Brin, Sergey},
  journal={Proceedings of ASIS, 1998},
  volume={98},
  pages={161--172},
  year={1998}
}

@inproceedings{louvain,
  title={Scalable community detection with the louvain algorithm},
  author={Que, Xinyu and Checconi, Fabio and Petrini, Fabrizio and Gunnels, John A},
  booktitle={2015 IEEE International Parallel and Distributed Processing Symposium},
  pages={28--37},
  year={2015},
  organization={IEEE}
}

@article{gae,
  title={Variational graph auto-encoders},
  author={Kipf, Thomas N and Welling, Max},
  journal={arXiv preprint arXiv:1611.07308},
  year={2016}
}

@article{wikipedia,
  title={Multi-scale attributed node embedding},
  author={Rozemberczki, Benedek and Allen, Carl and Sarkar, Rik},
  journal={Journal of Complex Networks},
  volume={9},
  number={2},
  pages={cnab014},
  year={2021},
  publisher={Oxford University Press}
}

@article{brody2021attentive,
  title={How attentive are graph attention networks?},
  author={Brody, Shaked and Alon, Uri and Yahav, Eran},
  journal={arXiv preprint arXiv:2105.14491},
  year={2021}
}

@inproceedings{pubmed,
  title={Revisiting semi-supervised learning with graph embeddings},
  author={Yang, Zhilin and Cohen, William and Salakhudinov, Ruslan},
  booktitle={International conference on machine learning},
  pages={40--48},
  year={2016},
  organization={PMLR}
}

@inproceedings{heterodata,
  title={Are we really making much progress? revisiting, benchmarking and refining heterogeneous graph neural networks},
  author={Lv, Qingsong and Ding, Ming and Liu, Qiang and Chen, Yuxiang and Feng, Wenzheng and He, Siming and Zhou, Chang and Jiang, Jianguo and Dong, Yuxiao and Tang, Jie},
  booktitle={Proceedings of the 27th ACM SIGKDD conference on knowledge discovery \& data mining},
  pages={1150--1160},
  year={2021}
}

@inproceedings{wang2019heterogeneous,
  title={Heterogeneous graph attention network},
  author={Wang, Xiao and Ji, Houye and Shi, Chuan and Wang, Bai and Ye, Yanfang and Cui, Peng and Yu, Philip S},
  booktitle={The world wide web conference},
  pages={2022--2032},
  year={2019}
}

@inproceedings{deepwalk,
  title={Deepwalk: Online learning of social representations},
  author={Perozzi, Bryan and Al-Rfou, Rami and Skiena, Steven},
  booktitle={Proceedings of the 20th ACM SIGKDD international conference on Knowledge discovery and data mining},
  pages={701--710},
  year={2014}
}

@inproceedings{metapath2vec,
  title={metapath2vec: Scalable representation learning for heterogeneous networks},
  author={Dong, Yuxiao and Chawla, Nitesh V and Swami, Ananthram},
  booktitle={Proceedings of the 23rd ACM SIGKDD international conference on knowledge discovery and data mining},
  pages={135--144},
  year={2017}
}

@article{gumbelsoftmax,
  title={Categorical reparameterization with gumbel-softmax},
  author={Jang, Eric and Gu, Shixiang and Poole, Ben},
  journal={arXiv preprint arXiv:1611.01144},
  year={2016}
}

@article{graph2feat,
  title={Graph2Feat: Inductive Link Prediction via Knowledge Distillation},
  author={Samy, Ahmed E and Kefato, Zekarias T and Girdzijauskas, {\v{S}}ar{\=u}nas},
  journal={Zenodo},
  doi={https://doi.org/10.1145/3543873.3587596},
  year={2023}
}

@article{zhu2021neural,
  title={Neural bellman-ford networks: A general graph neural network framework for link prediction},
  author={Zhu, Zhaocheng and Zhang, Zuobai and Xhonneux, Louis-Pascal and Tang, Jian},
  journal={Advances in Neural Information Processing Systems},
  volume={34},
  pages={29476--29490},
  year={2021}
}

@inproceedings{zhang2017weisfeiler,
  title={Weisfeiler-lehman neural machine for link prediction},
  author={Zhang, Muhan and Chen, Yixin},
  booktitle={Proceedings of the 23rd ACM SIGKDD international conference on knowledge discovery and data mining},
  pages={575--583},
  year={2017}
}

@article{zhang2018link,
  title={Link prediction based on graph neural networks},
  author={Zhang, Muhan and Chen, Yixin},
  journal={Advances in neural information processing systems},
  volume={31},
  year={2018}
}

@article{kipf2016semi,
  title={Semi-supervised classification with graph convolutional networks},
  author={Kipf, Thomas N and Welling, Max},
  journal={arXiv preprint arXiv:1609.02907},
  year={2016}
}

@article{hamilton2017inductive,
  title={Inductive representation learning on large graphs},
  author={Hamilton, Will and Ying, Zhitao and Leskovec, Jure},
  journal={Advances in neural information processing systems},
  volume={30},
  year={2017}
}

@article{zeng2019graphsaint,
  title={Graphsaint: Graph sampling based inductive learning method},
  author={Zeng, Hanqing and Zhou, Hongkuan and Srivastava, Ajitesh and Kannan, Rajgopal and Prasanna, Viktor},
  journal={arXiv preprint arXiv:1907.04931},
  year={2019}
}

@article{galkin2021nodepiece,
  title={Nodepiece: Compositional and parameter-efficient representations of large knowledge graphs},
  author={Galkin, Mikhail and Denis, Etienne and Wu, Jiapeng and Hamilton, William L},
  journal={arXiv preprint arXiv:2106.12144},
  year={2021}
}
\end{document}